\documentclass{article}
\usepackage{ijcai24}

\usepackage{times}
\usepackage{soul}
\usepackage{url}
\usepackage[hidelinks]{hyperref}
\usepackage[utf8]{inputenc}
\usepackage[small]{caption}
\usepackage{graphicx}
\usepackage{amsmath}
\usepackage{amsthm}
\usepackage{booktabs}
\usepackage{algorithm}
\usepackage{algorithmic}
\usepackage[switch]{lineno}

\usepackage[table,xcdraw]{xcolor}
\usepackage{xcolor}
\usepackage{multirow}
\usepackage{amssymb}

\title{FineDiffusion: Scaling up Diffusion Models for \\ Fine-grained Image Generation with 10,000 Classes}
\newcommand{\modelembedding}{TieredEmbedder}

\author{
Ziying Pan$^{1}$
\and
Kun Wang$^{4,5}$ \and
Gang Li$^{2,5,}$\thanks{ Corresponding author (e-mail: ucasligang@gmail.com,\\laiyx@xmu.edu.cn).} \and
Feihong He$^6$ \And
Yongxuan Lai$^{1,3,*}$ \\
\affiliations
$^1$School of Informatics, Xiamen University\\
$^2$Institute of Software, Chinese Academy of Sciences\\
$^3$School of Mathematics and Information Engineering, Longyan University\\
$^4$Shenyang Institute of Computing Technology of Chinese Academy of Sciences\\
$^5$University of Chinese Academy of Sciences\\
$^6$School of Computer Science and Technology, Soochow University\\
\emails
panziying@stu.xmu.edu.cn,
wk1360178179@gmail.com,
ucasligang@gmail.com,
18996341802@163.com,
laiyx@xmu.edu.cn
}

\begin{document}

\maketitle

 \begin{figure*}[t] 
    \centering
    \includegraphics[width=1\linewidth]{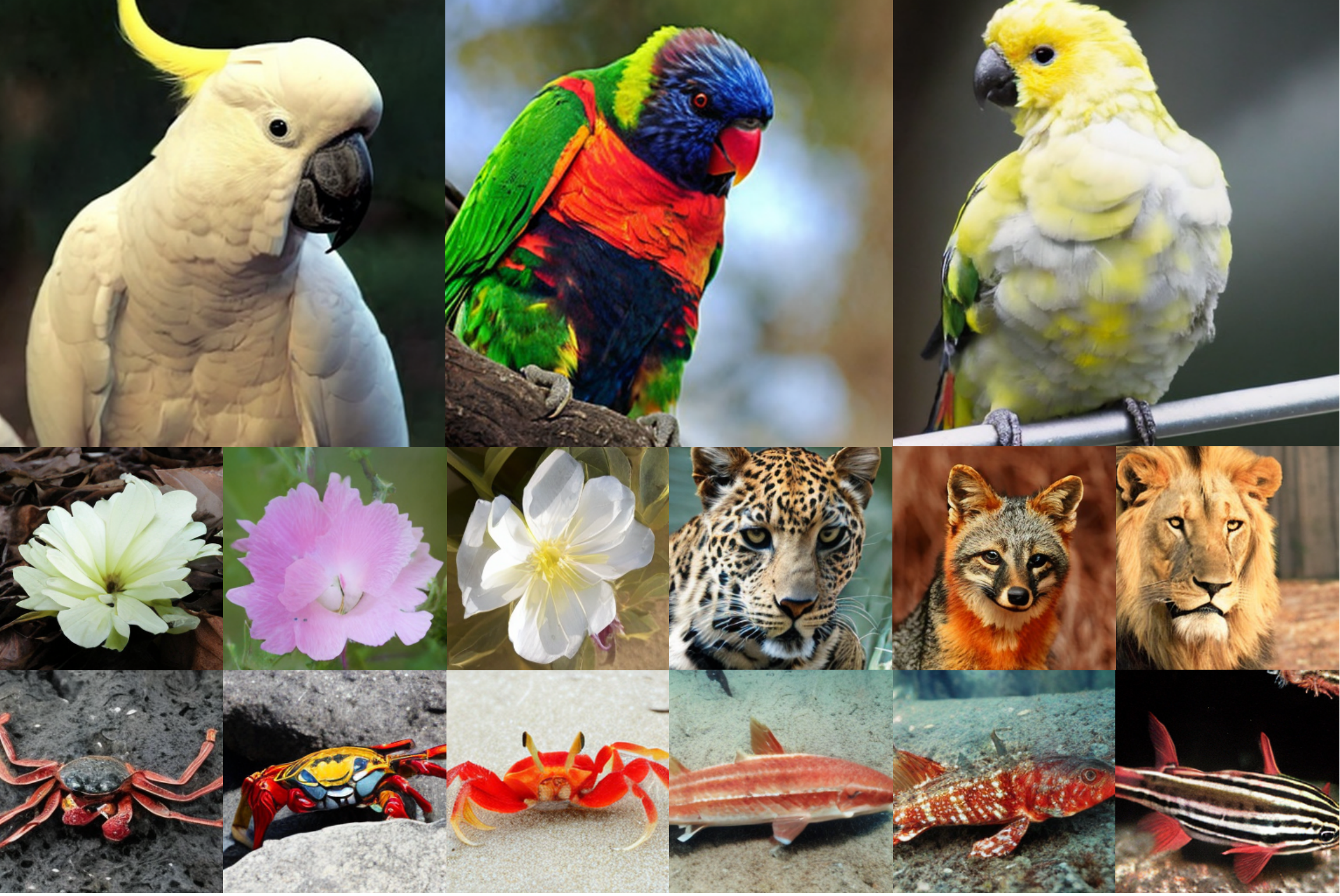} 
    \caption{Examples of images generated by the FineDiffusion model. The top row shows \textbf{Bird} images at a resolution of 512$\times$512 pixels, generated using the iNaturalist 2021 mini dataset. The bottom two rows are at a resolution of 256$\times$256 pixels, showing images of \textbf{Plant}, \textbf{Mammal}, \textbf{Arthropoda}, and \textbf{Ray-finned fish} generated from the same dataset. Each image is from a distinct fine-grained class, and every three images in the same row belong to one superclass. The generated results of fine-grained class images under the same superclass showcase the powerful fine-grained image generation capability of our method. }
    \label{fig:main_pic}
  \end{figure*}

\begin{abstract}
The class-conditional image generation based on diffusion models is renowned for generating high-quality and diverse images. However, most prior efforts focus on generating images for general categories, e.g., 1000 classes in ImageNet-1k. A more challenging task, large-scale fine-grained image generation, remains the boundary to explore. In this work, we present a parameter-efficient strategy, called \textit{FineDiffusion}, to fine-tune large pre-trained diffusion models scaling to large-scale fine-grained image generation with 10,000 categories. FineDiffusion significantly accelerates training and reduces storage overhead by only fine-tuning tiered class embedder, bias terms, and normalization layers' parameters. To further improve the image generation quality of fine-grained categories, we propose a novel sampling method for fine-grained image generation, which utilizes superclass-conditioned guidance, specifically tailored for fine-grained categories, to replace the conventional classifier-free guidance sampling. Compared to full fine-tuning, FineDiffusion achieves a remarkable 1.56$\times$ training speed-up and requires storing merely 1.77\% of the total model parameters, while achieving state-of-the-art FID of 9.776 on image generation of 10,000 classes. Extensive qualitative and quantitative experiments demonstrate the superiority of our method compared to other parameter-efficient fine-tuning methods. The code and more generated results are available at our project website: \href{https://finediffusion.github.io/}{https://finediffusion.github.io/}.

\end{abstract}

\section{Introduction}

Over the past decade, we have witnessed explosive growth in the realm of image generation due to the emergence of deep generative models. Recently, diffusion models~\cite{ho2020denoising_intro_art2,song2020denoising} have 
pushed back new frontiers of image generation, especially in text-to-image generation      ~\cite{nichol2021glide_intro_art8,rombach2022high_intro_art4,ramesh2022hierarchical_dalle2_intro_art9,saharia2022photorealistic_intro_art10} and class-conditional image generation~\cite{peebles2022scalable_intro_art5,gao2023masked_intro_art6}. 
These approaches outperform Generative Adversarial Networks~(GANs)~\cite{rw_dif_art1} in synthesizing high-quality and diverse images. 

Prior efforts for class-conditional image generation focus on generating images from general categories, e.g., the 1000 classes in ImageNet-1k~\cite{peebles2022scalable_intro_art5,gao2023masked_intro_art6,kim2022refining_intro_art13}. Meanwhile, training diffusion models for class-conditional image generation from scratch involves considerable computational costs, primarily due to their large parameter sizes and mode-covering behavior~\cite{rombach2022high_intro_art4}. Consider the recent Diffusion Transformer (DiT)~\cite{peebles2022scalable_intro_art5} as an illustration, specifically the DiT-XL/2 model, the largest variant in the DiT family renowned for its cutting-edge generative performance on the ImageNet class-conditional generation benchmark. To elaborate, DiT-XL/2 boasts 640 million parameters and involves computationally intensive training procedures. Training DiT-XL/2 on 256$\times$256 images demands 950 V100 GPU days (7 million iterations), and for 512$\times$512 images, it requires 1733 V100 GPU days (3 million iterations)~\cite{rw_dif_art13}. Relative to the image generation of general categories, fine-grained image generation poses a greater challenge, demanding models to intricately model subtle differences (e.g., feather texture of birds) in highly similar fine-grained categories. Therefore, training diffusion models for large-scale fine-grained class-conditional image generation from scratch requires greater computational resources and training iterations. We intend to develop an inexpensive method to fine-tune large pre-trained class-conditional diffusion models (e.g., DiT models) efficiently for large-scale fine-grained image generation.


Recent advancements in diffusion models, e.g., DiffFit~\cite{rw_dif_art13}, have demonstrated that satisfactory performance on downstream tasks can be attained by fine-tuning a small subset of parameters of the pre-trained model. In this paper, we further explore efficient fine-tuning strategies extending to large-scale fine-grained image generation tasks. We propose a novel fine-tuning method named FineDiffusion, capable of generating fine-grained images in downstream tasks by fine-tuning only the proposed TieredEmbedder, bias terms, and the scaling and offset terms of normalization layers. The proposed TieredEmbedder considers hierarchical class labels, including superclasses and subclasses, which are underutilized in image generation based on the diffusion model. 
Considering that large-scale fine-grained categories comprise various objects with noticeable appearance differences, knowing the superclass to which each class belongs can provide additional assistance for high-difficulty fine-grained image generation.
Moreover, we introduce a novel sampling method as an alternative to the conventional classifier-free guidance sampling for fine-grained image generation. Specifically, we combine noise estimation from the subclass-conditioned diffusion model and the jointly trained superclass-conditioned diffusion model. The guidance provided by the mixing weight facilitates the image generation toward specific subclasses within the superclass. By leveraging hierarchical category embedding information and employing fine-grained classifier-free guidance sampling, our FineDiffusion achieves finer control compared to unconditional diffusion guidance. It captures distribution differences among samples from various subclasses within the same superclass. The results of our fine-grained classifier-free guided sampling indicate that the diffusion model incorporating hierarchical information significantly improves the quality and diversity of generated images.
To the best of our knowledge, our work is the pioneering effort in utilizing diffusion-based techniques for fine-grained image generation encompassing 10,000 classes.
In performance metrics, FineDiffusion achieves an FID of 9.776 and an LPIPS of 0.721, surpassing the baseline algorithms substantially. 

\begin{figure}[t] 
    \centering
    \includegraphics[width=1\linewidth]{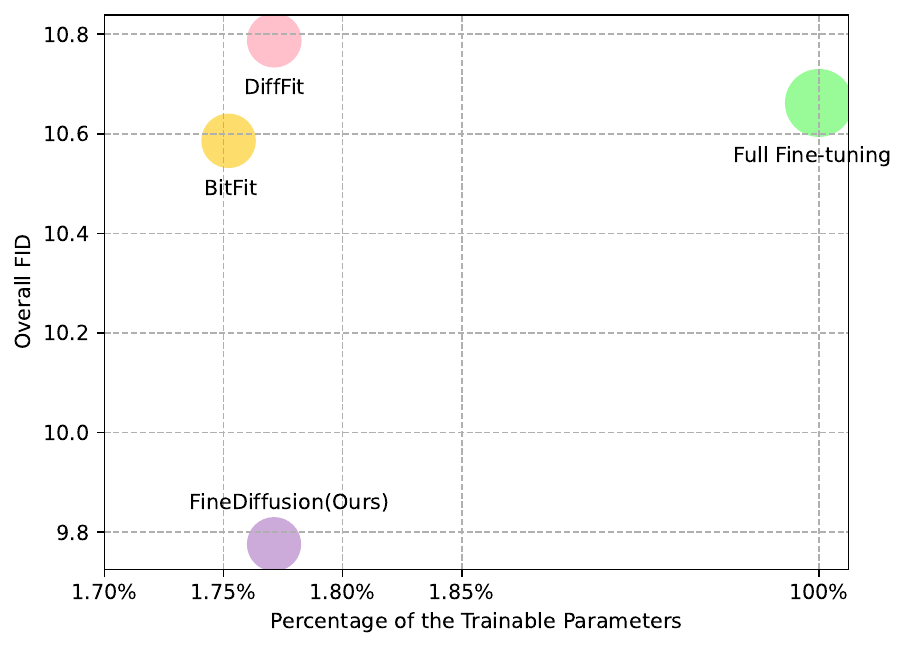} 
    \caption{The overall FID score is computed for the fine-tuned DiT model in the iNaturalist 2021 mini dataset. The size of each data point corresponds to the duration of training, with smaller points indicating faster training speed. FineDiffusion exhibits outstanding performance by attaining superior FID results, all the while demanding a reduced computational workload and fewer parameters.} 
    \label{fig:para_speed_fid_pic}
    \vspace{-4mm}
  \end{figure}

We summarize our contributions as follows:
\begin{itemize}
\item We introduce a novel parameter-efficient fine-tuning method called FineDiffusion for fine-grained image generation. We first propose a hierarchical class embedder called TieredEmbedder, which models the data distribution of both superclass and subclass samples.  By only fine-tuning proposed \modelembedding, bias terms, and the scaling and offset terms of normalization layers, our method significantly speeds up training while reducing model storage overhead (as shown in Figure~\ref{fig:para_speed_fid_pic} and Table~\ref{tab:para_speed}).

\item We introduce a fine-grained classifier-free guidance sampling method that enhances control over fine-grained image generation through conditional information from subclasses and superclasses.

\item Experimental results demonstrate that FineDiffusion outperforms state-of-the-art parameter-efficient fine-tuning methods in fine-grained image generation in terms of FID and LPIPS scores.

\end{itemize}

\section{Related Work}
\subsection{Diffusion Models} 

Diffusion models~\cite{sohl2015deep_intro_art1,ramesh2022hierarchical_dalle2_intro_art9,saharia2022photorealistic_intro_art10,li20223ddesigner,he2024freestyle} have become the most prevalent generative models, offering more stable training and generating higher-quality visual content compared to earlier mainstream models such as GANs~\cite{rw_dif_art1}, VAEs~\cite{rw_dif_art2}, and Flow-based models~\cite{rw_dif_art3}.
Denoising diffusion probabilistic models (DDPMs)~\cite{ho2020denoising_intro_art2} architecture, based on U-Net~\cite{ronneberger2015unet}, has propelled the diffusion model to significant success in image generation. Following this, improvements in sampling techniques have further enhanced DDPM's performance~\cite{song2020denoising,rw_dif_art9,rw_dif_art9}. Classifier-free guidance~\cite{rw_dif_art9} proposes a diffusion guidance method without additional classifiers by jointly training conditional and unconditional diffusion models. For its impressive image generation results, numbers of models have adopted classifier-free guidance, making significant strides in the image generation field~\cite{nichol2021glide_intro_art8,ramesh2022hierarchical_dalle2_intro_art9,gafni2022make_cl1,mokady2023null_c3}. 
However, classifier-free guidance inadequately considers the correlations among different classes in fine-grained fields. In light of this, we explore a fine-grained classifier-free guidance approach that effectively utilizes the label hierarchy structure to improve fine-grained image generation. 
Additionally, recent works~\cite{rw_dif_art11,peebles2022scalable_intro_art5} have explored the integration of Transformer architectures into diffusion models. DiT models~\cite{peebles2022scalable_intro_art5} have garnered significant attention due to their superior scalability and simplicity, inspiring numerous subsequent works~\cite{rw_dif_art13,mo2024dit,he2023cartoondiff,lu2024fit}. We follow the DiT~\cite{peebles2022scalable_intro_art5} paradigm and further fine-tune the pre-trained models for fine-grained image generation.

\subsection{Parameter-efficient Fine-tuning} 
The traditional full fine-tuning approach~\cite{rw_ft_art1_howard2018universal,rw_ft_art2_dai2015semi}, which requires tuning all parameters, is resource-intensive and time-consuming.
To solve this issue, parameter-efficient fine-tuning has gained momentum.
Some researches introduce additional parameters, such as plug-and-play modules or the incorporation of side network structures, for fine-tuning the small subset of parameters introduced additionally~\cite{rw_ft_art3_houlsby2019parameter,rw_ft_art4_hu2021lora,chen2022adaptformer,he2022sparseadapter}. On the flip side, some researches have seamlessly extended successful prompt tuning techniques from the field of NLP (Natural Language Processing) to CV (Computer Vision)~\cite{rw_ft_prot_art2,rw_ft_prot_art3,rw_ft_prot_art4,rao2022denseclip}. Other researches explore the fine-tuning of only crucial parameters within the model to enhance the efficiency of the fine-tuning process~\cite{zaken2021bitfit,zhao2020masking,xu2021raise_part1}. There are also hybrid fine-tuning methods, e.g., DiffFit~\cite{rw_dif_art13}, which fine-tune only specific parameters related to bias, class embedding, normalization, and additional introduced scale factors.
However, DiffFit is a general image generation model fine-tuning strategy lacking specific enhancements tailored for fine-grained image generation. In contrast, our approach possesses specific innovations in this regard.

\section{Methodology}

\begin{figure}[t] 
    \centering
    \includegraphics[width=\linewidth]{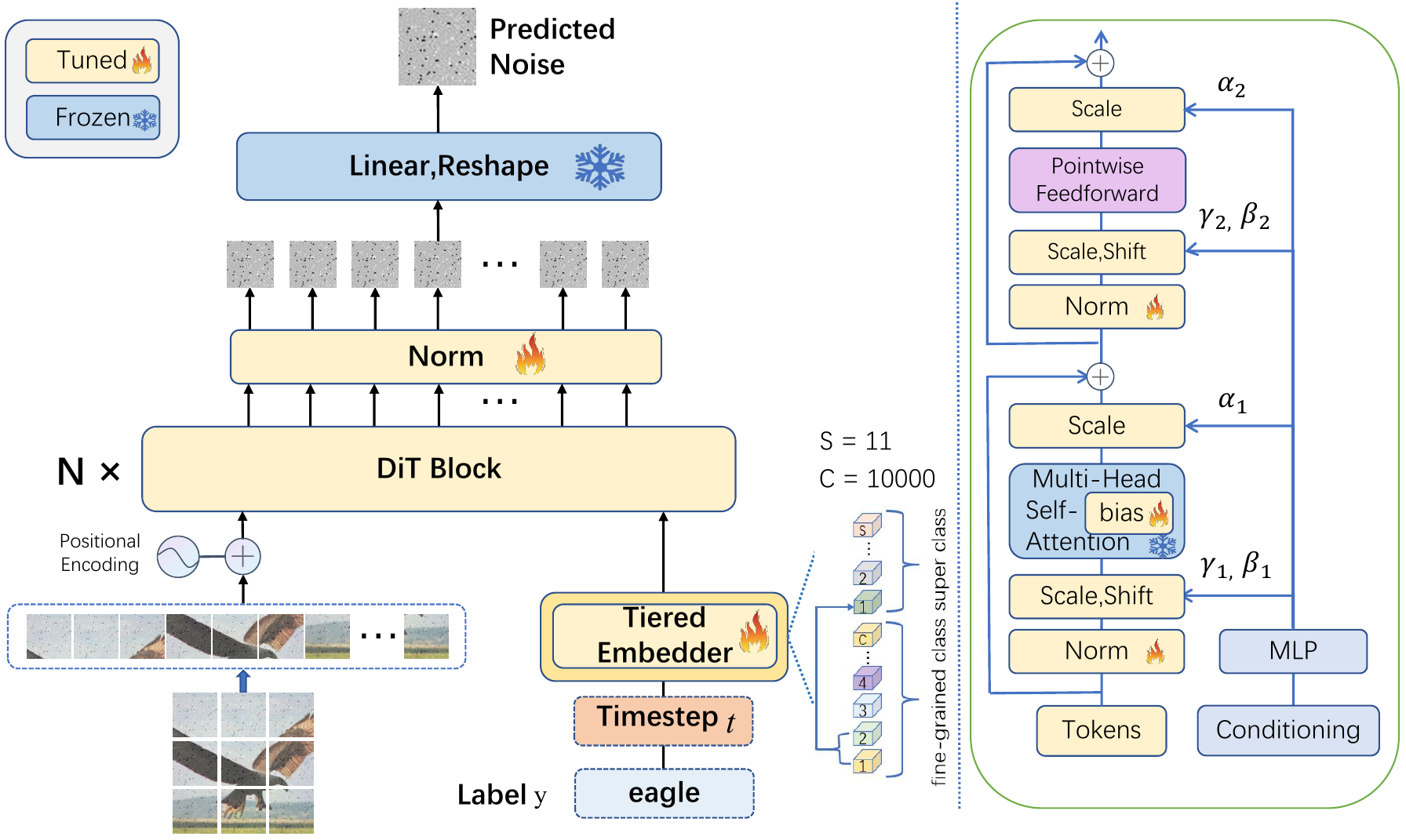} 
    \caption{The proposed FineDiffusion method involves the preservation of the majority of parameters within the pre-trained DiT model. We introduce a specialized TieredEmbedder optimized for generating fine-grained categories. Notably, we exclusively fine-tune the tiered label embedding component, bias terms, and normalization terms, and this fine-tuning process affects merely 1.77\% of the pre-trained model's parameters. This strategic approach showcases an effective means of achieving efficient parameter fine-tuning. (Zoom-in for the best view.)}
    \label{fig:overview}
    \vspace{-3mm}
\end{figure}


\subsection{Preliminaries}
\paragraph{Diffusion Models.}
Diffusion models are currently one of the most mainstream generative models, consisting of two processes: forward noise injection and backward denoising.
In the forward process, Gaussian noise is added to the original image $x_{0}$ to produce the noisy image $x_t$, where $t\sim[1, T]$. The image $x_t$ is obtained by the formula $x_{t}=\sqrt{\bar{\alpha}_{t}} x_{0}+\sqrt{1-\bar{\alpha}_{t}} \epsilon, \text { where } \epsilon \sim \mathcal{N}(0, \mathbf{I}) \text{ and } \bar{\alpha}_t=\prod_{t=1}^t\alpha_t$,
$\alpha_t$ is the pre-defined hyperparameter related to the time step t.
The backward process is defined as a Markov chain which beginning with $x_T \sim \mathcal{N}(0, \mathbf{I}) $, $x_{t-1}$ can be determined by $x_t$ according to the following formula:
$p_{\theta}\left(x_{t-1} \mid x_{t}\right)=\mathcal{N}\left(x_{t-1} ; \mu_{\theta}\left(x_{t}, t\right), \sigma_{t}^{2} \mathbf{I}\right)$. After T iterations of denoising, the clean image $x_0$ is obtained.






\paragraph{Diffusion Transformers (DiT).}

DiT~\cite{peebles2022scalable_intro_art5} is a recent notable approach that formulates a diffusion model using Transformer architecture. Meanwhile, it also adopts the idea of Latent Diffusion Models (LDM)~\cite{rombach2022high_intro_art4} to perform image denoising within the latent space.
DiT consists of two parts for image generation given sample $x$:
(1) Comprising an encoder ($E$) and a decoder ($D$), the Variational Autoencoder (VAE) operates such that the latent code is represented as $z=E(x)$, and the reconstructed data is denoted as $\hat{x} =D(z)$. (2) A latent diffusion transformer incorporating patchify, sequential DiT blocks, and depatchify operations. The combination of embedded labels and iteration time step embeddings \(t\) serves as a condition for the training of the DiT model, with the latent used as input.

\paragraph{Classifier-free Guidance (CFG).} CFG~\cite{rw_dif_art9} is a sampling strategy independent of the classifier. Distinct from the classifier-guidance strategy, it significantly reduces training complexity while ensuring high-quality image generation.
During training, conditional and unconditional diffusion models are jointly trained, being parameterized using a score estimator $\epsilon_{\theta}$.
During sampling, for the given condition $c$, the conditional diffusion models follow the formula:
$\hat{\epsilon}_{\theta}\left(x_{t} \mid c \right)=\epsilon_{\theta}\left(x_{t} \mid \varnothing\right)+\omega  \cdot\left(\epsilon_{\theta}\left(x_{t} \mid c \right)-\epsilon_{\theta}\left(x_{t} \mid \varnothing\right)\right)$, where $\omega$ is the guidance scale to adjust the condition-relatedness of the results, and the null label $\varnothing$ input denotes the estimation of the unconditional model.
CFG has been widely applied in diffusion models~\cite{nichol2021glide_intro_art8,ramesh2022hierarchical_dalle2_intro_art9,peebles2022scalable_intro_art5} because of the exceptional performance.
\subsection{FineDiffusion}
\subsubsection{3.2.1 Parameter-efficient Fine-tuning}
In this part, we provide technical details on how FineDiffusion fine-tunes the DiT model for fine-grained image generation.
The overview of our FineDiffusion is depicted in Figure~\ref{fig:overview}.
In the DiT model, the label embedding segment encodes class labels to be utilized as conditions for class-conditional image generation. For image generation of a specific class, the corresponding class embedding vector is concatenated with the noised latent image, serving as the input for the sampling process.
For fine-grained image generation, the inclusion of superclass information facilitates a more precise capture of category-specific features, thereby enhancing the overall quality of generated images.
We propose a hierarchical class label encoding strategy that concurrently encodes superclass and subclass labels. In the DiT model, the label embedding segment accepts class labels as input and outputs a table with fixed-dimensional embedding vectors. We introduce embeddings representing the superclass into the label embedding segment and train superclass embeddings together with other subclass embeddings, to minimize parameter introduction. And the novel embedder module responsible for encoding hierarchical class labels is named TieredEmbedder.
After training the proposed TieredEmbedder, both superclass and subclass labels are input to the diffusion model for guiding fine-grained image generation.
To further improve fine-tuning effectiveness, we consider adjusting parameters beyond the TieredEmbedder.
Inspired by DiffFit~\cite{rw_dif_art13}, we simultaneously fine-tune bias and normalization terms along with the tiered embedder to learn the global dataset's distribution characteristics.
Through these designs, our proposed FineDiffusion aims to deepen the model's understanding of image content. By integrating higher-level semantic structures and learning dataset-wide, superclass, and subclass information, it enhances the conditional generation images' quality.

During fine-tuning, the majority of parameters in the pre-trained DiT model are are frozen, and updates are applied only to TieredEmbedder, bias terms, and normalization terms. 
FineDiffusion requires fine-tuning only 1.77\% of the parameters in the DiT-XL/2 pre-trained model, achieving a training speed-up of 1.56$\times$ compared to full fine-tuning. Simultaneously, as an efficient parameter fine-tuning method, the proposed approach enhances adaptability for fine-grained image generation tasks while effectively mitigating the risk of catastrophic forgetting. For each dataset fine-tuning, it only needs to store the small subset of parameters undergoing fine-tuning, which dramatically reducing the model's storage requirements. 

\subsubsection{3.2.2 Fine-grained Classifier-free Guidance}
Classifier-free guidance has supplanted classifier guided sampling to achieve a trade-off between the diversity and quality of sampling. While reducing training complexity, this approach secures high-quality image generation. Currently, classifier-free guidance is widely utilized in various diffusion models.
We attempt to introduce hierarchical category label information because knowing which superclass each category belongs to is helpful for capturing intricate details and enhancing the overall quality of the generated images. 
Therefore, we introduce a fine-grained classifier-free guidance sampling method, yielding improved results under specific conditions. Integrated with TieredEmbedder, this approach leverages superclass conditional information to enhance control over generated images.
Conventional classifier-free guidance involves training both conditional and unconditional generative models. In our approach, a superclass conditional generative model is introduced to replace the unconditional model.
During training, we use the same neural network to parameterize both the conditional model and the superclass conditional model, leveraging a score estimator $\epsilon_{\theta}$ for optimization guidance.
Subclass labels $c$ are probabilistically replaced with their corresponding superclass labels $c_{p}$ to achieve the learning of superclass embeddings.
During sampling, trained superclass and subclass embeddings are input into the DiT network to obtain predictive noise. Our method combines conditional scores estimation $\epsilon_{\theta}\left(x_{t} \mid c\right)$ and superclass conditional scores estimation $\epsilon_{\theta}\left(x_{t} \mid c_{p}\right)$ to guide specific category image generation, with the specific formula provided below:

\begin{equation}
\resizebox{0.9\linewidth}{!}{$
    \hat{\epsilon}_{\theta}\left(x_{t} \mid c \right)=\epsilon_{\theta}\left(x_{t} \mid c_{p}\right)+\omega  \cdot\left(\epsilon_{\theta}\left(x_{t} \mid c \right)-\epsilon_{\theta}\left(x_{t} \mid c_{p}\right)\right),
      $}
\end{equation}
where $\omega$ is the guidance scale, and $\hat{\epsilon}_{\theta}$ is the modified score.

By thoroughly exploiting the distributional characteristics of superclasses and subclasses, our sampling method achieves a finer control than classifier-free guidance. We can capture commonalities among subclasses within the same superclass and simultaneously capture distinct distributions between different superclasses, thereby making the learned distributions of each category more closely approximate real-world situations.


\begin{table*}[t]
\label{tab:metrics_comparison}
\centering
\renewcommand\arraystretch{1.3}
\setlength{\tabcolsep}{0.055cm}

\scalebox{1}{
\begin{tabular}{ccccccccccccc}
\cline{1-13}
                         & \multicolumn{2}{c}{Plants}            & \multicolumn{2}{c}{Insects}       & \multicolumn{2}{c}{Birds}         & \multicolumn{2}{c}{Fungi}         & \multicolumn{2}{c}{Reptiles}      & \multicolumn{2}{c}{Mammals}         \\ \cline{2-13}
\multirow{-2}{*}{Method}  & FID$\downarrow$   & LPIPS$\uparrow$   & FID$\downarrow$ & LPIPS$\uparrow$ & FID$\downarrow$ & LPIPS$\uparrow$ & FID$\downarrow$ & LPIPS$\uparrow$ & FID$\downarrow$ & LPIPS$\uparrow$ & FID$\downarrow$ & LPIPS$\uparrow$   \\ \cline{1-13}
Full Fine-tuning          & 13.200           & 0.671             & 18.477          & 0.733           & 21.554          & 0.720           & 27.259          & 0.657           & 37.828          & 0.686           & 38.061          & 0.715             \\
BitFit                   & \textbf{12.025}            & 0.670             & 19.328          & 0.740           & 23.677          & 0.723           & 29.589          & 0.662           & 41.845          & 0.683           & 39.918          & 0.714             \\
DiffFit                   & 12.166                         & 0.668             & 19.809          & 0.738           & 24.063          & 0.721           & 30.000          & 0.659           & 42.657          & 0.688           & 40.243          & \textbf{0.716}             \\
\rowcolor{gray!20}
FineDiffusion\textbf{(ours)}      & 15.752            & \textbf{0.693}             & \textbf{13.871}          & \textbf{0.741}           & \textbf{14.713}          & \textbf{0.727}           & \textbf{26.805}          & \textbf{0.695}           & \textbf{25.266}          & \textbf{0.709}           & \textbf{34.567}          & \textbf{0.716}       
\\ \cline{1-13}
                         & \multicolumn{2}{c}{Ray-finned Fishes} & \multicolumn{2}{c}{Amphibians}    & \multicolumn{2}{c}{Mollusks}      & \multicolumn{2}{c}{Arachnids}     & \multicolumn{2}{c}{Animalia}      & \multicolumn{2}{c}{Overall}           \\ \cline{2-13}
\multirow{-2}{*}{Method} & FID$\downarrow$   & LPIPS$\uparrow$   & FID$\downarrow$ & LPIPS$\uparrow$ & FID$\downarrow$ & LPIPS$\uparrow$ & FID$\downarrow$ & LPIPS$\uparrow$ & FID$\downarrow$ & LPIPS$\uparrow$ & FID$\downarrow$ & LPIPS$\uparrow$  \\ \cline{1-13}
Full Fine-tuning          & 35.613            & 0.702             & 55.077          & 0.713           & 44.075          & 0.717           & 45.722          & 0.732           & 50.038          & 0.698           & 10.662          & 0.708            \\
BitFit                   & 41.265            & 0.706             & 66.261          & 0.714           & 44.594          & 0.725           & 45.287          & 0.744           & 51.904          & 0.700           & 10.586          & 0.710             \\
DiffFit                  & 44.340            & 0.698             & 68.030          & 0.714           & 44.845          & 0.731           & 45.865          & 0.742           & 52.059          & 0.697           & 10.788          & 0.709            \\
\rowcolor{gray!20} 
FineDiffusion\textbf{(ours)}     &  \textbf{30.348}            & \textbf{0.716}                     &     \textbf{33.827}         & \textbf{0.717}           & \textbf{38.590}          & \textbf{0.735}           & \textbf{33.214}          & \textbf{0.751}           & \textbf{47.353}           & \textbf{0.704}           & \textbf{9.776}  & \textbf{0.721} \\
\cline{1-13}
\end{tabular}
}
\caption{FID and LPIPS metrics comparison on with existing methods full fine-tuning, BitFit~\protect\cite{zaken2021bitfit} and DiffFit~\protect\cite{rw_dif_art13} on 11 categories of iNaturalist 2021 mini dataset image synthesis.}
\label{tab:main_fid_lpips_result}
\vspace{-3mm}
\end{table*}

\begin{table}[t]
\label{tab:para_train_speed}
\centering
\renewcommand\arraystretch{1}
\setlength{\tabcolsep}{0.04cm}
\begin{tabular}{cccc}
\hline
Method            & Params.(M)    & 
\begin{tabular}[c]{@{}c@{}}Training Cost\\ (GPU days)\end{tabular} & \begin{tabular}[c]{@{}c@{}}Training\\ Time\end{tabular} \\ \hline
Full Fine-tuning     & 685.2(100\%)  & 15.11                                                               & 1$\times$                                                      \\
BitFit              & 12.01(1.75\%) & 9.59                                                               & 0.635$\times$                                                  \\
DiffFit             & 12.14(1.77\%) & 9.65                                                               & 0.638$\times$                                                  \\
\rowcolor{gray!20}
FineDiffusion\textbf{(ours)} & 12.15(1.77\%)   & 9.69                                                      &          0.642$\times$                                                  \\ \hline
\end{tabular}
\caption{Comparison of the number of parameters and training overhead on iNaturalist 2021 mini image synthesis.}\label{tab:para_speed}
\end{table}

\begin{table}[t]
\label{tab:vegfru_metrics}
\centering
\renewcommand\arraystretch{1}
\setlength{\tabcolsep}{0.55cm}
\begin{tabular}{cccc}
\hline
Method            & FID$\downarrow$     & LPIPS $\uparrow$ \\ \hline
FineDiffusion-256& 9.776                                                       &          0.721                                                  \\
FineDiffusion-512& \textbf{9.490}  & \textbf{0.723}                                                \\ \hline
\end{tabular}
\caption{Comparison of the overall FID and overall LPIPS metrics for FineDiffusion fine-tuning on the 256$\times$256 and 512$\times$512 resolutions of the DiT/XL-2 model .}\label{tab:main_fid_lpips_result_512}
 \vspace{-5mm}
\end{table}

\begin{table}[t]
\centering
\renewcommand\arraystretch{1}
\setlength{\tabcolsep}{0.55cm}
\begin{tabular}{cccc}
\hline
Method            & FID$\downarrow$     & LPIPS $\uparrow$ \\ \hline
Full Fine-tuning     & 13.034                                                                & 0.651                                                      \\
BitFit              & 15.022                                                                & 0.654                                                 \\
DiffFit             & 15.068                                                                & 0.653                                                  \\
\rowcolor{gray!20}
FineDiffusion\textbf{(ours)} & \textbf{12.382}                                                       &          \textbf{0.667}                                                  \\ \hline
\end{tabular}
\caption{Comparison of the overall FID and overall LPIPS metrics on VegFru dataset.}\label{tab:vegfru_result}
\end{table}



\begin{figure}[t] 
\centering
\includegraphics[width=1\linewidth]{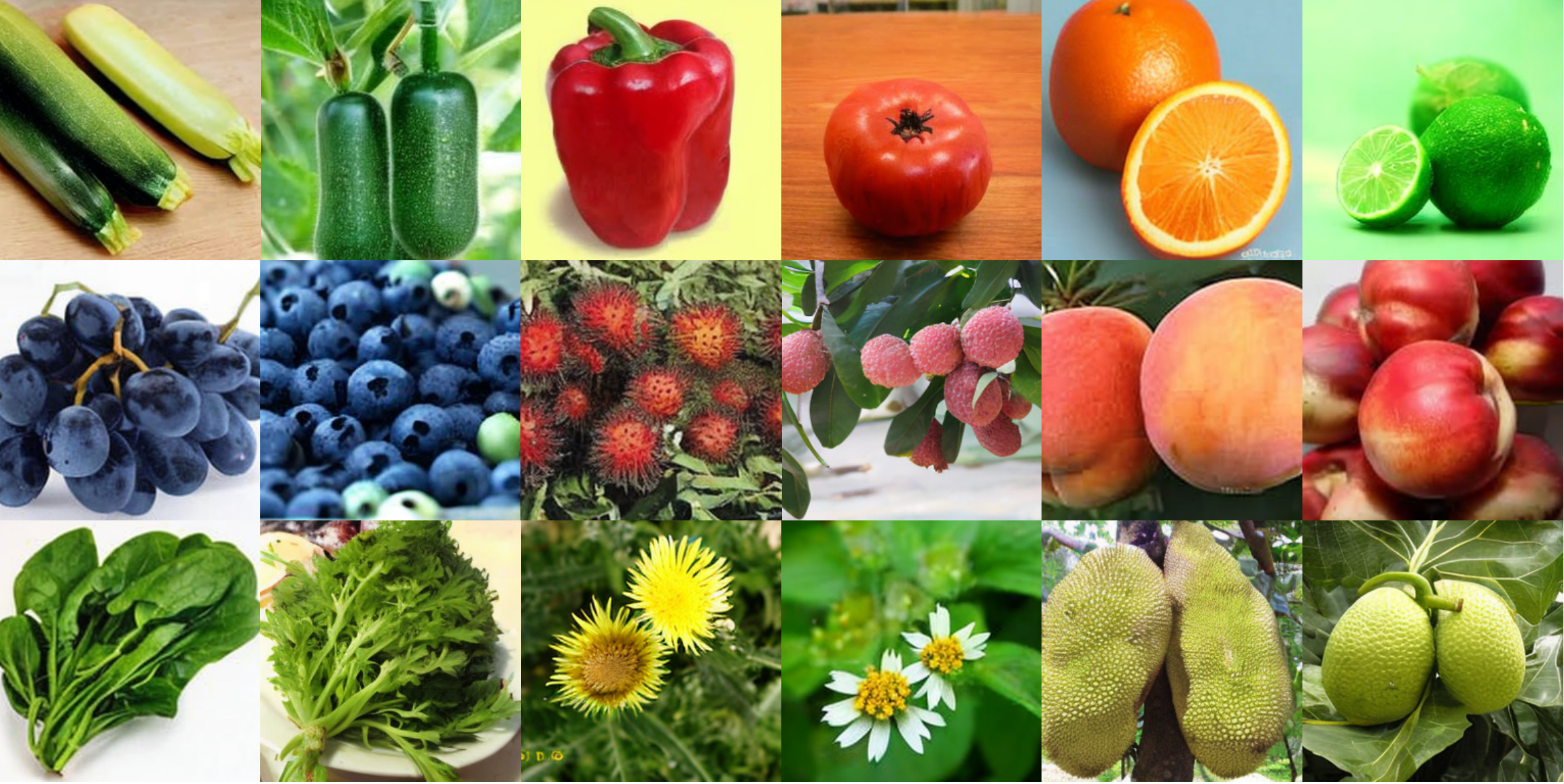} 
\caption{Images generated by the FineDiffusion model after fine-tuning on the VegFru dataset. In each row, every pair of images belongs to the same superclass, namely: Melon, Eggplant, Citrus fruit, Litchies, Berry fruit, Drupe, Green-leafy vegetable, Wild vegetable and Collective fruit. Our approach effectively generates subclasses with visually similar features within the same superclass.} 
\vspace{-3mm}
\label{fig:vegfru_pic}
\end{figure}

\section{Experiments}
\subsection{Datasets}
\textbf{iNaturalist 2021.} The iNaturalist dataset~\cite{van2018inaturalist_intro}, in its 2021 version, represents a comprehensive real-world collection featuring 10,000 distinct species across 11 supercategories. In our experiments, we employed the mini version of the iNaturalist 2021 training dataset, consisting of 500,000 training samples. The official validation set, containing 100,000 images, is utilized for metric testing.

\noindent \textbf{VegFru.} The VegFru dataset~\cite{hou2017vegfru} comprises 292 fine-grained categories of vegetables and fruits, organized into 25 superclasses. Its training set contains 29,200 images, and the test set contains 14,600 images. We utilize the dataset splits provided by the official source.

\begin{figure}[ht] 
\centering
\includegraphics[width=1.0\linewidth]{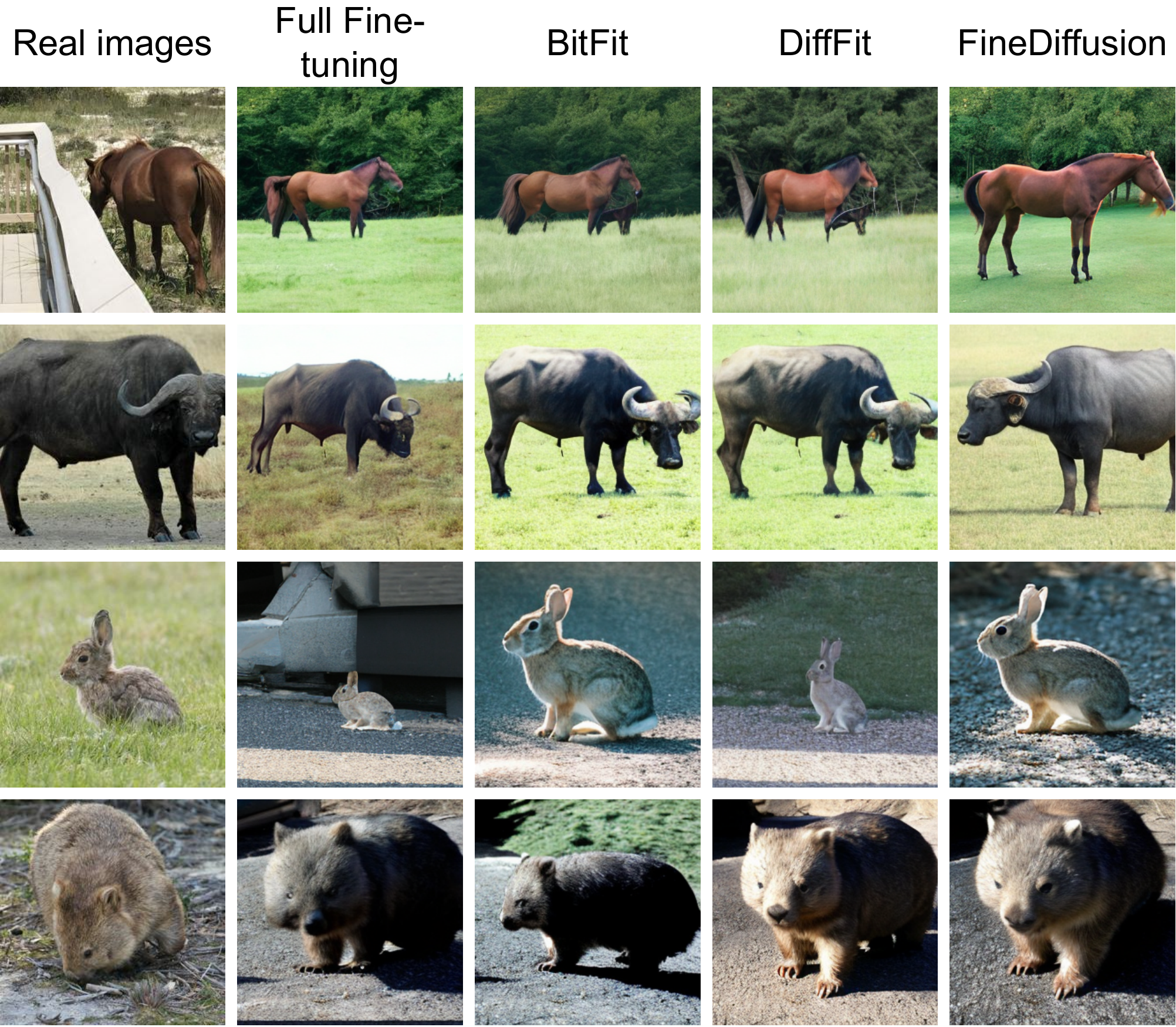} 
\caption{
Comparison of generated results for several mammalian categories across different methods. Each method is assigned the same class label input and the same random seed for sampling. The authentic images of real species are also presented. This comparison underscores FineDiffusion's capability to generate species images aligned with actual categories, and these images are more photorealistic compared to other methods. (Zoom-in for the best view.)
} 
\vspace{-5mm}
\label{fig:same_seed_animals}
\end{figure}

\subsection{Implementation Details}
We utilize the pre-trained DiT-XL/2 model as the backbone of our model, \emph{i.e.,} the model size is XL, patchified with the size of 2. 
The original DiT model is trained on ImageNet-1k 256$\times$256. 
The fine-tuning training process involves leveraging V100 GPUs, training 200K iterations on the iNaturalist 2021 mini training set, and training 100K iterations on the VegFru dataset. The model is optimized with AdamW~\cite{loshchilov2017decoupled_adamw} optimizer and the learning rate is 1e-4. 
The sampling procedure adopts the classifier-free guidance with a scale of 4.0 and uses 250 steps for the sampling process.
When examining the overall results, the generated images are based on randomly selected sample seeds. 
To evaluate the convergence speed of each algorithm, a uniform set of fixed sample seeds was employed across distinct algorithms, ensuring controlled variables. 

\subsection{Evalution Metrics}
We utilize the Fréchet Inception Distance (FID)~\cite{experiments_evalution_r1} and Learned Perceptual Image Patch Similarity (LPIPS)~\cite{experiments_evalution_r2} metrics to assess the model's performance. 
FID quantifies the differences in mean and covariance of features between real and generated images in the feature space, and it reflects their similarity through traces. 
A smaller FID value indicates a higher degree of similarity between the two image sets. LPIPS is primarily employed to measure the similarity between image samples, with higher values indicating greater perceptual dissimilarity between samples. This implies that the generated fine-grained images exhibit significant diversity, which correlates with better performance.


\begin{figure}[ht] 
\centering
\includegraphics[width=1\linewidth]{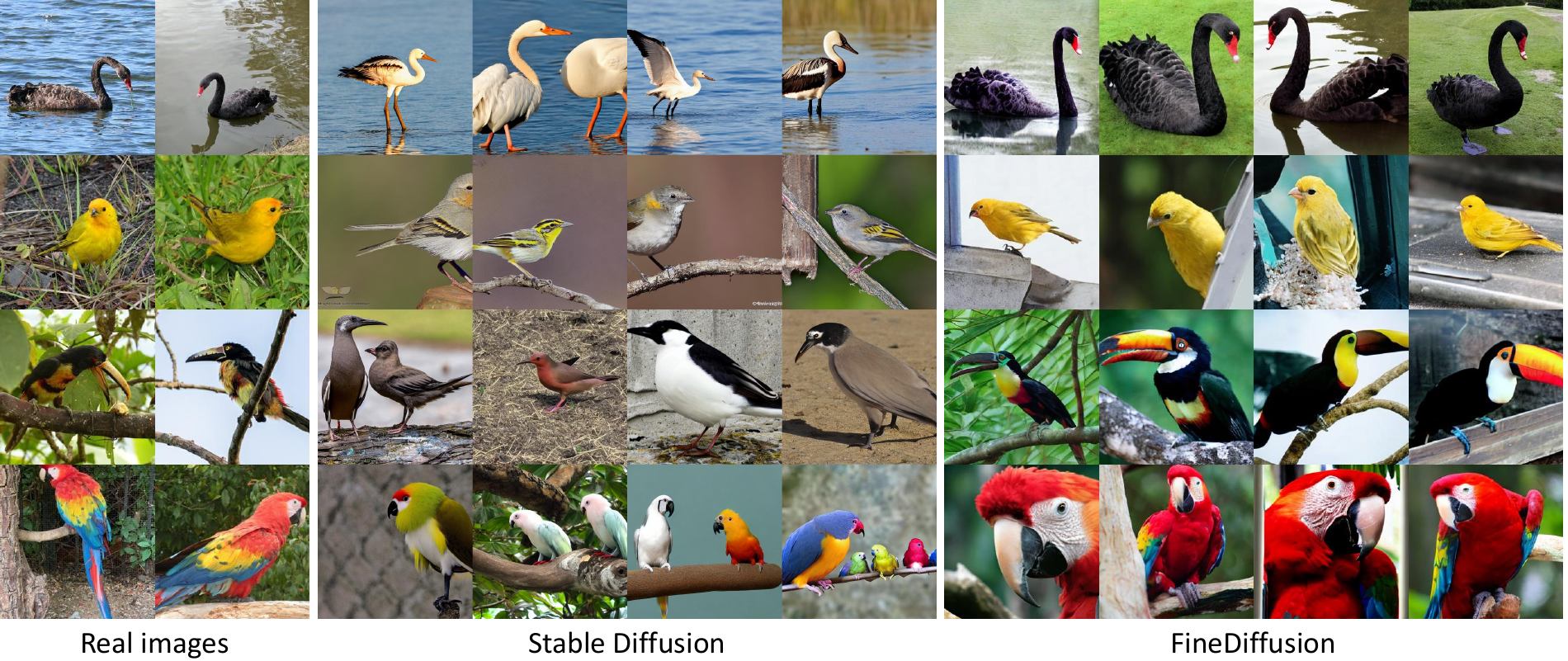} 
\caption{Comparison with stable diffusion in the generation of various fine-grained bird categories. The first two columns depict real images, while the images in the subsequent columns, organized by rows, showcase generated images for each fine-grained category. FineDiffusion achieves superior accuracy in generating avian species that closely resemble those depicted in real images. (Zoom-in for the best view.)} 
\vspace{-5mm}
\label{fig:sd_comparison}
\end{figure}

\begin{figure*}[t] 
    \centering
    \includegraphics[width=1\linewidth]{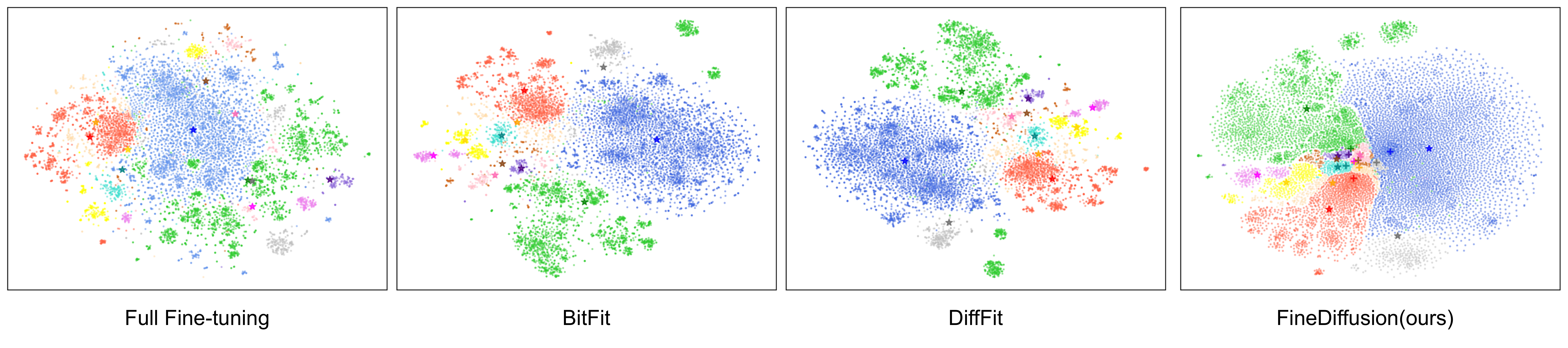} 
    \vspace{-3mm}
    \caption{The t-SNE visualization of label embedding distributions obtained from different fine-tuning methods. Diverse colors represent unique superclasses within iNaturalist 2021, where ``$\bullet$'' represents subclasses in each superclass.  ``$\star$'' represents the average embeddings of subclasses belonging to the same superclass, while the ``$\scalebox{0.7}{+}$'' represents the embeddings of the superclass. (Zoom-in for the best view.)} 
    \label{fig:tsne_comparison}
    \vspace{-3mm}
\end{figure*}

\subsection{Quantitative Comparison}
\textbf{Results on iNaturalist.} The outcomes are presented in Tables~\ref{tab:main_fid_lpips_result}, \ref{tab:para_speed} and \ref{tab:main_fid_lpips_result_512}, accompanied by generated samples illustrated in Figure~\ref{fig:main_pic}. Through quantitative comparison, we evaluate our algorithm alongside three comparative algorithms: full fine-tuning, BitFit (fine-tuning only the bias terms), and DiffFit (fine-tuning bias terms, normalization layers, and additional scaling terms).
In the generation of 11 super categories, the proposed FineDiffusion fine-tuning based on 256$\times$256 resolution model achieved an exceptional overall FID of 9.776 with only 1.77\% parameter fine-tuning, as shown in Table~\ref{tab:main_fid_lpips_result} and~\ref{tab:para_speed}.
Significantly, FineDiffusion exhibits superior performance, achieving the best FID scores in 10 out of the 11 categories. In comparison, three baseline algorithms exhibit similar FID scores. The full fine-tuning approach shows relatively good FID performance in each category, demonstrating its robust performance as a baseline method. However, it's important to note that full fine-tuning with 100\% parameters leads to significant computational resource wastage and large storage cost for more downstream tasks.
BitFit obtained the second-best overall FID, trailing behind FineDiffusion, through 1.75\% parameter fine-tuning,  showing its parameter efficiency.
In contrast, DiffFit exhibited subpar performance compared to several algorithms. This could be attributed to DiffFit being more suited for generating images of generalized categories, or there might be a need to explore parameter settings more compatible with DiffFit.
The exceptional FID scores of the FineDiffusion model highlight the efficacy of fine-tuning tiered embeddings and guided sampling utilizing superclass information. This strategy effectively preserves the knowledge acquired by the pre-trained model, facilitating successful transfer to diverse fine-grained species.
In terms of the LPIPS metric, the FineDiffusion model consistently outperforms other models in all categories, as well as maintaining the highest overall LPIPS score. This suggests that our algorithm not only enhances the quality of fine-grained images generation but also exhibits a high level of diversity.

Table~\ref{tab:para_speed} demonstrates that the FineDiffusion model requires only fine-tuning a minimal amount of parameters, effectively conserving model storage space.
It fine-tunes only a few more parameters than other efficient fine-tuning algorithms, while resulting in a significant improvement in the generated results. 
This not only reduces storage costs but also accelerates training speed, making it more resource-efficient compared to full fine-tuning, achieving an astonishing 1.56$\times$ training speed-up. The accelerated training speed of FineDiffusion bears substantial implications for practical applications. Encouraged by the stellar performance of FineDiffusion on the 256$\times$256 resolution DiT model, we proceeded to fine-tune a 512$\times$512 resolution DiT/XL-2 model to acquire images of heightened quality. The experimental results are shown in Table \ref{tab:main_fid_lpips_result_512}, revealing commendable FID and LPIPS values across various categories. The aggregate FID reach a noteworthy 9.490, while the LPIPS score attain 0.723. The 512-resolution model demonstrates a further enhancement over the performance of the 256-resolution model.



\noindent \textbf{Results on VegFru.} Additionally, we assess our method on the VegFru dataset, and the results are presented in Table~\ref{tab:vegfru_result}, accompanied by generated samples in Figure~\ref{fig:vegfru_pic}. For brevity, we offer overall results instead of an exhaustive list of categories of 25 superclasses.
Notably, FineDiffusion surpasses all counterparts with a markedly  FID of 12.382 and LPIPS of 0.667, suggesting superior performance in capturing image distribution nuances. In terms of FID, it achieves a reduction of approximately 5.0\% compared to full fine-tuning, 17.6\% compared to BitFit, and 17.8\% compared to DiffFit. 
This indicates the stable applicability of our methodology across diverse fine-grained image scenarios, offering valuable supplementation to the experiments conducted on the iNaturalist dataset.


\subsection{Qualitative Comparison}

In Figure~\ref{fig:same_seed_animals}, we present the results of mammalian class generation using the FineDiffusion method and juxtapose them with several baseline fine-tuning algorithms. 
In comparison to other methods, our model excels in accurately capturing the inherent features of each species, producing life-like depictions of animals with realistic lighting and shadow effects. The resulting images exhibit a pronounced sense of authenticity and exceptional quality.
Significantly, it is noteworthy that in contrast to other algorithms which entail more extensive parameter fine-tuning, our approach overcomes problems such as morphological distortions or image blurriness. This observation implies that our fine-tuning approach maximally retains the valuable information of the pre-trained model, and efficiently adapts to fine-grained generation tasks, yielding visually coherent and accurate generative outcomes.

We also qualitatively compare with state-of-the-art text-guided image generation models, e.g., Stable Diffusion~\cite{rombach2022high_intro_art4}. For our approach, images are generated based on category labels, whereas Stable Diffusion generates images using the prompt phrase ``a photo of xxx'' (where ``xxx" indicates the species name).
The comparison results are shown in Figure~\ref{fig:sd_comparison}, revealing that FineDiffusion excels in generating accurate images for specific fine-grained species. In contrast, images generated by Stable Diffusion exhibit a certain degree of similarity under various fine-grained category prompts. 


\subsection{Visualization of Class Embeddings}
We utilize the t-SNE~\cite{TSNE_experiments_evalution_r3} technique to visualize high-dimensional label embedding vectors in a two-dimensional context. Employing eleven distinct colors, we portrayed 10,000 classes in iNaturalist, revealing the distribution of embeddings across eleven superclasses, as shown in Figure~\ref{fig:tsne_comparison}. 
Examining the class distributions, we observe that in the FineDiffusion approach, points within the same superclass tend to cluster together, while points from different superclasses remain separate. The points within each superclass are relatively dispersed, and the majority of mean points are positioned at the center of their respective classes. The embeddings of superclasses also tend to lie among their respective subclasses, which indicates the effectiveness of superclass embedding learning.
In contrast, the full fine-tuning method shows less separation among superclass embeddings, with points from different superclasses mixing. Comparatively, BitFit and DiffFit effectively distinguish superclasses, while there is some clustering among points within the same superclass, which is not as well-dispersed as FineDiffusion.
The t-SNE results indicate the superior fine-tuning effectiveness of FineDiffusion on label embeddings, enhancing discriminability among distinct classes and even within subclasses of the same supercategory. This is crucial for generating fine-grained images across a large scale of 10,000 classes.

\section{Conclusion}
In this work, we make the first attempt to scale up diffusion models for fine-grained image generation with 10,000 classes.
We introduce FineDiffusion, an efficient parameter-tuning approach that fine-tunes key components, including tiered label embeddings, bias terms, and normalization terms of pre-trained models. Our method substantially reduces both training and storage overheads. Furthermore, we introduce a fine-grained classifier-free guidance sampling technique, harnessing hierarchical data label information to effectively enhance the performance of fine-grained image generation. 
Adequate qualitative and quantitative results demonstrate the superiority of our method compared to the other methods.


\bibliographystyle{named}
\bibliography{ijcai24}

\end{document}